\documentclass{article}
\usepackage{adjustbox}
\PassOptionsToPackage{numbers}{natbib}
\usepackage{natbib}

     \usepackage[preprint]{neurips_2022}

\usepackage[utf8]{inputenc} 
\usepackage[T1]{fontenc}    
\usepackage{hyperref}       
\usepackage{url}            
\usepackage{booktabs}       
\usepackage{amsfonts}       
\usepackage{nicefrac}       
\usepackage{microtype}      
\usepackage{subfig}
\usepackage[capitalize]{cleveref}
\usepackage{graphicx}
\usepackage{booktabs,makecell}
\usepackage{lipsum}

\usepackage{caption,stackengine}
\title{RainUNet for Super-Resolution Rain Movie Prediction under Spatio-temporal Shifts}

 \author{%
   Jinyoung Park \\
   KAIST \\
   \texttt{jinyoungpark@kaist.ac.kr} \\
   \And
   Minseok Son  \\
   KAIST \\
   \texttt{ksos104@kaist.ac.kr} \\
   \AND
   Seungju Cho \\
   KAIST \\
   \texttt{joyga@kaist.ac.kr} \\
   \And
   Inyoung Lee \\
   KAIST \\
   \texttt{inzero24@kaist.ac.kr} \\
   \And
   Changick Kim \\
   KAIST \\
   \texttt{changick@kaist.ac.kr} \\
 }

\begin{document}

\maketitle

 \begin{abstract}
 This paper presents a solution to the Weather4cast 2022 Challenge Stage 2. 
 The goal of the challenge is to forecast future high-resolution rainfall events obtained from ground radar using low-resolution multiband satellite images. 
 We suggest a solution that performs data preprocessing appropriate to the challenge and then predicts rainfall movies using a novel RainUNet.
 RainUNet is a hierarchical U-shaped network with temporal-wise separable block (TS block) using a decoupled large kernel 3D convolution to improve the prediction performance. 
  Various evaluation metrics show that our solution is effective compared to the baseline method.
  The source codes are available at \url{https://github.com/jinyxp/Weather4cast-2022}.
 \end{abstract}

\section{Introduction}
With the recent rapid climate change, accurately predicting the weather in the near future has become important \cite{veillette2020sevir, klocek2021ms}.
Accordingly, several deep learning-based efficient weather forecasting methods have been proposed over several years \cite{klocek2021ms, woo2017operational, trebing2021smaat, bai2022rainformer}.
In Weather4cast 2022\cite{9672063,10.1145/3459637.3482044}, super resolution rain movie prediction under spatio-temporal shifts is presented as a challenging task.
The data provided for the Weather4cast 2022 competition have been obtained from a satellite and a ground radar in three European regions over two years, which is taken at 15 minutes intervals.
In each region, the satellite captures 11-band spectral videos with three modalities (infrared, visible, and water vapor).
The 11 channels comprise seven infrared (IR), two visible (VIS), and two water vapor (WV) channels are used as inputs.
The one-channel videos are captured from the ground radar and used as outputs, representing pixel-wise rain rates with 1 and 0 for rainy or non-rainy, respectively. 
The core challenge of this competition is predicting the one-channel rain rate sequences for the next 8 hours in the middle area corresponding to 2km$\times$2km using the 11-band spectral sequences for the past hour corresponding to a 12km$\times$12km region.
Additionally, a transfer challenge across space and time aims to verify the transferability of the proposed model. 
In constructing a dataset for the above scenario, four consecutive 11-band spectral frames constitute the input sequence, and 32 consecutive rain rate future frames constitute the ground-truth sequence for comparison with the prediction results.
All frames in the sequences are $252\times252$ pixels in size.
This paper proposes a simple yet novel hierarchical U-shaped RainUNet for the Weather4cast 2022 competition. The proposed model shows competitive performances for the core and transfer challenges.

\section{Method}
\subsection{Data Preprocessing}
\begin{figure}[h]
\centering
\captionsetup[subfigure]{labelformat=empty}
\begin{adjustbox}{minipage=\linewidth,scale=0.75}
\subfloat{ 
    \includegraphics[width=0.24\textwidth]{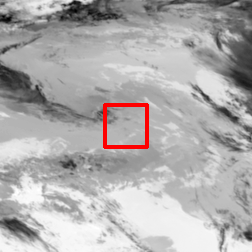}
}
\subfloat{
    \includegraphics[width=0.24\textwidth]{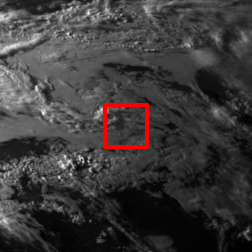} 
}
\subfloat{ 
    \includegraphics[width=0.24\textwidth]{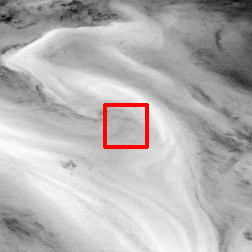}
}
\subfloat{
    \includegraphics[width=0.24\textwidth]{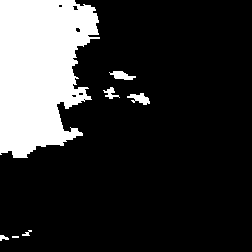}
} \\ 
\subfloat[IR]{ 
    \includegraphics[width=0.24\textwidth]{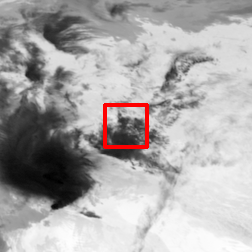}
}
\subfloat[VIS]{
    \includegraphics[width=0.24\textwidth]{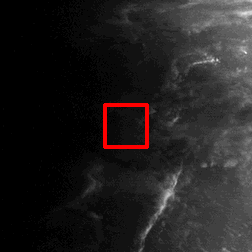} 
}
\subfloat[WV]{ 
    \includegraphics[width=0.24\textwidth]{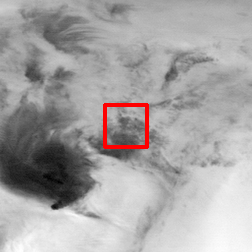}
}
\subfloat[target]{
    \includegraphics[width=0.24\textwidth]{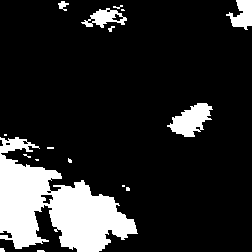}
}

\label{fig:channels}
  \end{adjustbox}
  \caption{
  Example of three different sensing modalities (IR, VIS, and WV) and the target image visualization corresponding to the same time.
   Target area is highlighted
in red box.
  Here, we randomly select \texttt{IR\_087}, \texttt{VIS\_006}, \texttt{WV\_062} images for IR, VIS and WV.}
  \label{fig:dataset}
\end{figure}

\textbf{Sensing Modality Selection.}
The input data of the Weather4cast 2022 challenge comprise three different sensing modalities: seven channels of IR, two channels of VIS, and two channels of WV.
Visualization of each channel type and target images is shown in \cref{fig:dataset}.
We mainly use IR channel, which occupies the largest proportion of the satellite images, and then conduct a toy experiment by adding VIS and WV channel types to determine which types to use. 
We train the base model 3D U-Net with IR, IR + VIS, IR + WV, and IR + VIS + WV and summarize the results in \Cref{tab:channnels}.
Overall, IR + VIS produces the best performances, considering IoU, precision, and accuracy.
Therefore, we use a total of nine channels comprising IR and VIS for our method.

\vspace{-0.5cm}

\begin{table}[ht!]
  \caption{
  Performances based on sensing modality selection. 
  The best in bold.}
  \vspace{0.2cm}
  \centering
    
  \begin{tabular}{lll|c c c c c}
    \toprule
    \multicolumn{3}{c|}{Sensing modality} &
    \multicolumn{5}{c}{Metrics}                  \\
    \cmidrule(r){1-3} \cmidrule(r){4-8}
    IR&VIS&WV & IoU & Precision & Recall & Accuracy  & F1\\
    \midrule
    \checkmark& & & 0.0615 & 0.0638 &  0.4260 &0.7201  & 0.1148 \\
    \checkmark&\checkmark& & \textbf{0.1026} & \textbf{0.1152} & 0.4627 & \textbf{0.7774} & \textbf{0.1837} \\
    \checkmark& & \checkmark&  0.0597 & 0.0752 & 0.3982 & 0.7577 & 0.1116 \\
    \checkmark&\checkmark& \checkmark& 0.0961 & 0.1041 & \textbf{0.5496} & 0.7227 & 0.1732 \\
    \bottomrule
  \end{tabular} 
  \label{tab:channnels}
\end{table}

\textbf{Data Cleansing.}
The data sequences used in the experiment include unbalanced meteorological situations, although images are obtained at various times in various places.
Generally, since non-rainy situations occur more often than rainy situations, the entire data are biased toward non-rainy situations.
This trains the model with a bias for non-rainy situations; thus, we filter the data to reduce the weather bias of the entire data. 
One data sequence has four past frames for each selected modality and 32 ground-truth future frames for the rain rate.
We count whole positive pixels in 32 ground-truth future frames, and if the values are less than 100, the sequences are regarded as non-rainy situations and removed.
Among the total 228,928 sequences, 69,693 non-rainy sequences are excluded from the training phase.

\newpage
\textbf{Center Crop.}
Both the input satellite radiance and output OPERA ground-radar rain rates are given for $252\times252$ pixel patches; however, the spatial resolution of the satellite images is approximately six times lower than the resolution of the ground radar. 
Simply, the area in the output frames corresponds to the central $42\times42$ pixel area in the input frame.
That is, the predictive model exploits $252\times252$ pixels to predict future frames with a size of $42\times42$ pixels.
In this case, the input image contains spatial information for a too wide area, it could rather hinder the learning of the model.
To prevent such adverse effects, we crop the center of the input frame to three times the size of the target region and resize it to $252\times252$ pixels for training.

\begin{figure}[t]
      
      \centering
      \resizebox{13.5cm}{!}{
      \includegraphics{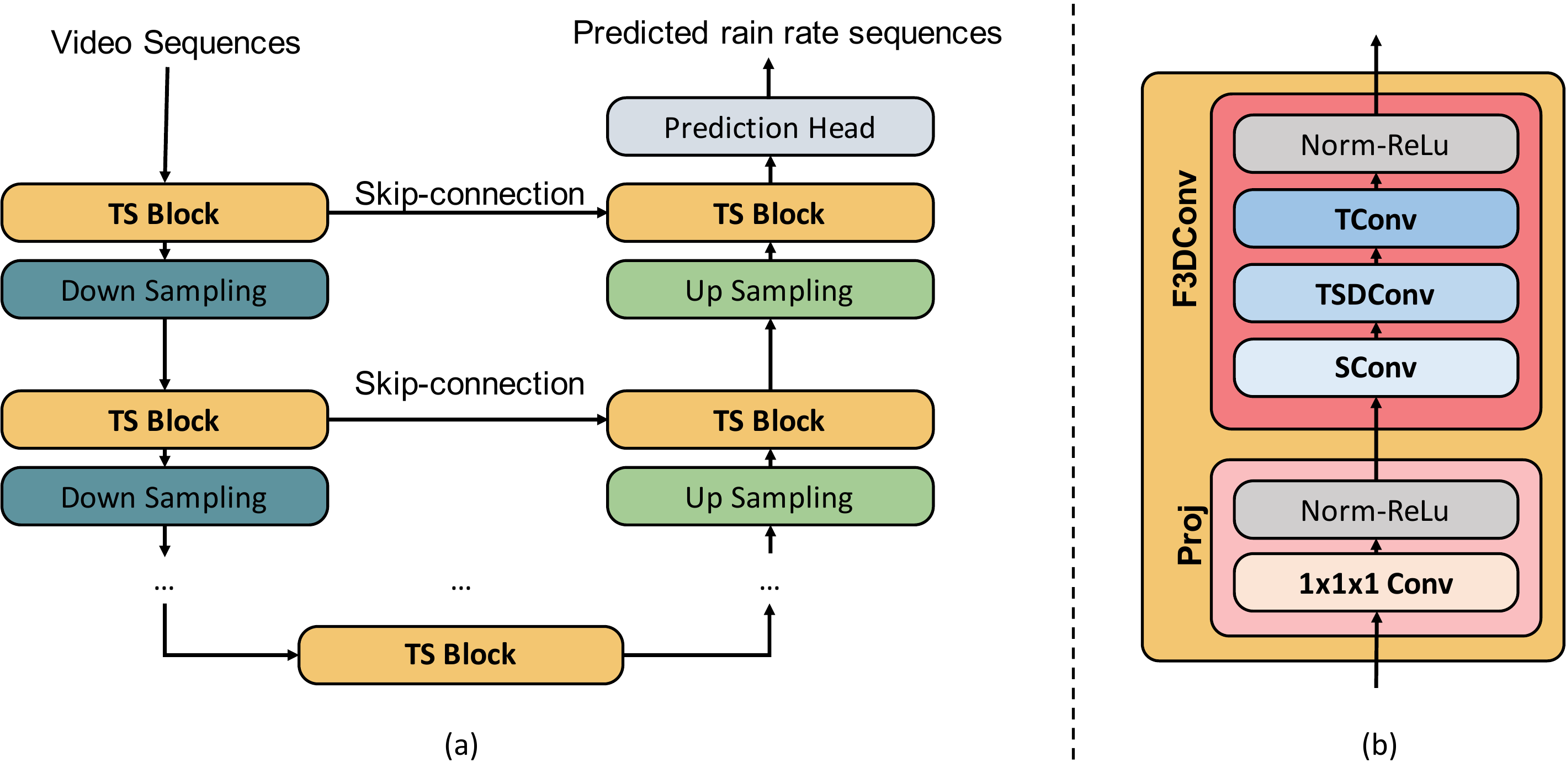}
      }
      \caption{(a) Overall structure of RainUNet. (b) The detailed structure of our TS block.}
      \label{fig:main}
\end{figure}

\subsection{RainUNet}
\textbf{Overall Pipeline.}
As shown in \cref{fig:main} (a), the overall structure of the proposed RainUNet is a hierarchical U-shaped network. 
Following the 3D U-shaped structures~\cite{cciccek20163d}, intermediate features are passed through skip-connections between encoder and decoder.
The encoder extracts spatiotemporal patterns through $K$ stages, and each stage contains a proposed a temporal-wise separable block (TS block) and a down-sampling layer. 
The TS block takes advantage of both CNN and Transformer, which has low computational cost compared to the attention mechanism while capturing long-range dependencies.
In the downsampling layer, we downsample the feature maps using a 3D max pooling operation with kernel size 2.

The proposed decoder progressively integrates spatiotemporal information for $K$ stages to predict future rain rates.
Each decoder block comprises an upsampling layer and an TS block.
here, we use a 3D transposed convolution for the upsampling, and this layer doubles the size of the feature maps while halving the channels. 
The skip-connection then concatenates the processed features with the encoder features of the corresponding stage. 
Subsequently, the TS block is applied to learn to predict future rain rates by integrating extracted spatiotemporal information.
After the final $K$ decoder stages, a $ 1\times1\times1 $ convolution layer is used to reduce the channels and predict future rain rates. In our experiments, we empirically set $K=5$.
We train RainUNet using the dice loss (DL)~\cite{sudre2017generalised}:
\begin{equation}
DL = 1-\frac{2\sum_{1=1}^{N}p_ig_i}{\sum_{i=1}^{N}p_i^2+\sum_{i=1}^{N}g_i^2} ,
\end{equation}
where $p_i \in P$ is the predicted probability of the $i-th$ pixel, and $g_i \in  G$ is the ground truth of the $i-th$ pixel of a total $N(=252\times252)$ pixels.
The $P = [0, 1]^N$ and $G = \{0, 1\}^N$ denote predicted probability maps and ground truths, respectively.


\clearpage
\textbf{TS Block.} To capture the spatiotemporal pattern of meteorological features, we propose the TS block with two submodules: the $\texttt{LinearProjection }$module and the factorized 3D convolution ($\texttt{F3DConv}$) module. Here, we use a $1 \times 1 \times 1$ convolution for $\texttt{LinearProjection}$. The $\texttt{F3DConv}$ characterizes richer interframe correlations within an intraframe global region rather than only the local neighboring region in consecutive frames. TS block yields significant gains in model performances by exploiting spatiotemporal features suitable for weather forecasting.

\textbf{F3DConv.} We construct $\texttt{F3DConv}$  as shown in \cref{fig:main} (b).  
We decompose spatiotemporal modeling into two separate steps: spatial and temporal modeling. 
To this end, we disentangle the $t \times d \times d$ 3D convolution into a $1 \times d \times d$ spatial convolution ($\mathrm{SConv}$) and a $t \times 1 \times 1$ temporal convolution ($\mathrm{TConv}$), where $d$ denotes the spatial kernel size of width and height, and $t$ denotes the temporal extent of the filter. 
Moreover, the spatial convolution is followed by the temporal convolution to enable interframe temporal modeling based on the existing video backbones \cite{qiu2017learning,tran2018closer,qiu2019learning,qiu2022optimization}.
However, due to the locality of the spatial convolution, its receptive field is limited, which makes it incapable of investigating intraframe context, which is an excellent clue to future weather prediction. 
To alleviate this issue, we inject a spatially dilated convolution and the temporal wise dilated convolution ($\mathrm{TSDConv}$) between the 2D spatial convolution and the 1D temporal convolution.
$\mathrm{TSDConv}$ generates long-range spatial features, and enables the model to focus on the surrounding area required to predict the next frames.
In this work, we adopt $1\times3\times3$ spatial convolution, $1\times7\times7$ temporal-wise dilated convolution with dilation 3 and $3\times1\times1$ temporal convolution.

The computation of a TS block is represented as follows:
\begin{equation}
\vspace{0.1cm}
    Z_l =\mathtt{\texttt{F3DConv}}(\mathtt{Proj}(Z_{l-1})),
\end{equation}
\begin{equation}
    \mathtt{Proj}
    =\sigma(\mathrm{Norm}(\mathrm{Conv}_{1\times1\times1}(F))).
\end{equation}
\begin{equation}
    \mathtt{\texttt{F3DConv}} =\sigma(\mathrm{Norm}(\mathrm{TConv}(\mathrm{TSDConv}(\mathrm{SConv(F)}))).
\vspace{0.1cm}
\end{equation}

Here, given the previous stage features $Z_{l-1}$ as an input of the ${l-th}$ stage block,  $Z_l$ is the output of the $\texttt{F3DConv}$ module.
$\mathrm{Norm}$ and $\sigma$ denote a group normalization \cite{wu2018group} and the ReLU activation function, respectively. 
We note that our factorization is closely related to a large kernel convolution \cite{guo2022visual}, which was proposed to decompose standard 2D convolution into spatial and channel axes and successfully increase the receptive field to the image segmentation task.
However, the large kernel convolution is based on a 2D convolution and requires additional considerations for treating the temporal information when extended to space-time data.
We propose instead decomposing 3D convolution into spatial and time axes and forcing the 3D convolution into discrete spatial and temporal components to not only increase interframe correlation.

\textbf{Stochastic Weight Averaging.}
Stochastic weight averaging (SWA) is well known procedure that leads to better generalization and wider optima \cite{izmailov2018averaging}.
SWA averages the weights of the models to take advantage of the model ensemble, which improves generalization performance without increasing computational complexity.
Since the model generalization is related to transfer learning, we use the SWA-applied model for the transfer challenge task.

\section{Experiments}
\subsection{Experimental Settings}
For a fair comparison, all models are trained on the training set covering seven regions and 2 years and report the average scores of five metrics on the all core validation set: IoU, precision, recall, accuracy, and F1 score.
Additionally, we provide the test leader board score for stage 2 (test score) on the test datasets.
We apply center crop (\texttt{Crop}) and data cleansing (\texttt{DC}) for data preprocessing. 
During training, we employ AdamW \cite{loshchilov2017decoupled} with a betas=$(0.9, 0.999)$, and a minibatch size of 80 to optimize the models. 
The initial learning rate is set to $1 \times 10^{-3}$. 
All models are trained for 20 epochs from scratch on four NVIDIA RTX 3090 GPUs. 
Note that SWA is only used for the transfer challenge task to boost generalization ability during training.
\clearpage
\subsection{Experimental Results}

\begin{table}[ht]
\vspace{-0.4cm}
  \caption{
  Comparing our RainUnet with the baseline. RainUnet performs better than the baseline approach on validation dataset and test dataset.
  The best results are shown in bold.}
  \label{tab:main_table}
  \centering
  \resizebox{13.8cm}{!}{
  \begin{tabular}{l c c| c c | c c c c c | c}
    \toprule
    \multicolumn{1}{c}{} &
    \multicolumn{2}{c|}{Preprocessing } &
    \multicolumn{7}{c|}{Metrics} &
    \multicolumn{1}{c}{Test} \\
    \cmidrule(r){2-10}
    Model & \texttt{DC} & \texttt{Crop} & \#Param.(M) &Training time &IoU&Precision&Recall&Accuracy&F1&Score\\
    \midrule
    3D U-Net & & & 22.6& 38& 0.2134& 0.2538& 0.4488& 0.8434& 0.3408&0.2256\\
    RainUNet & & & 49.6& 45& 0.2310& 0.2954& 0.5084& 0.8775& 0.3616&0.2455\\
    RainUNet &\checkmark && 49.6& 34& 0.2358& 0.3004& 0.5053& 0.8786& 0.3663&0.2403\\
    RainUNet &&\checkmark & 49.6& 49& 0.2473& \textbf{0.3227}& 0.5123& 0.8843& 0.3848&0.2648\\
    RainUNet &\checkmark &\checkmark & 49.6& 37& \textbf{0.2508}& 0.3221& \textbf{0.5325}& \textbf{0.8857}& \textbf{0.3874}& \textbf{0.2685}\\
    \bottomrule
  \end{tabular} 
  }
  \label{tab:aug}
\end{table}

\begin{table}[ht!]
\vspace{-0.4cm}
  \caption{Ablation study on RainUnet structure. 3D U-Net is a default structure and replaces with RainUnet's encoder and decoder.
  Best in bold and second in underline.}
  \label{tab:encdec}
  \centering
  \resizebox{13.8cm}{!}{
  \begin{tabular}{cc|c|c c c c c| c}
    \toprule
    \multicolumn{2}{c}{ } &
    \multicolumn{6}{c|}{Metrics} &
    \multicolumn{1}{c}{Test}\\
    \cmidrule(r){3-8} 
    
    \texttt{Enc}&\texttt{Dec}&\#Param. (M)&IoU&Precision&Recall&Accuracy&F1 & Score\\
    \midrule
    &                    & 22.6& 0.2134& 0.2538& 0.4488& 0.8434&0.3408&0.2256\\
    \checkmark&          & 44.1& \underline{0.2485}    & 0.3192& \underline{0.5189}& \underline{0.8858}& \underline{0.3856}&0.2470\\
    &\checkmark          & 28.1&     0.2421            & \textbf{0.3229}& 0.4962& \textbf{0.8876}&  0.3782&\underline{0.2673}\\
    \checkmark&\checkmark& 49.6& \textbf{0.2508}       & \underline{0.3221}& \textbf{0.5325}& 0.8857&\textbf{0.3874}&\textbf{0.2685}\\
    \bottomrule
  \end{tabular} }
\end{table}

The results of the baseline 3D U-Net and our methods are shown in \Cref{tab:aug}.
Evidently, RainUNet is superior in all metrics compared to 3D U-Net. 
Additionally, the overall performance on various metrics is boosted when the \texttt{DC} and \texttt{Crop} are combined with our method.
Furthermore, we achieve a slightly shorter training time through \texttt{DC}, making our method efficient and effective.

\Cref{tab:encdec} summarizes the results when the encoder and decoder parts are replaced by our proposed method, respectively.
The best performance is shown in the IoU, recall, and F1 score when the TS blocks are applied to both the encoder and decoder.
Regarding precision and accuracy, since no significant difference from the second best is observed, we adopt a model in which the TS blocks are applied to both the encoder and decoder. 

\begin{figure}[h!]
\vspace{-0.4cm}
\begin{minipage}[b]{0.64\textwidth}
\includegraphics[scale=0.39]{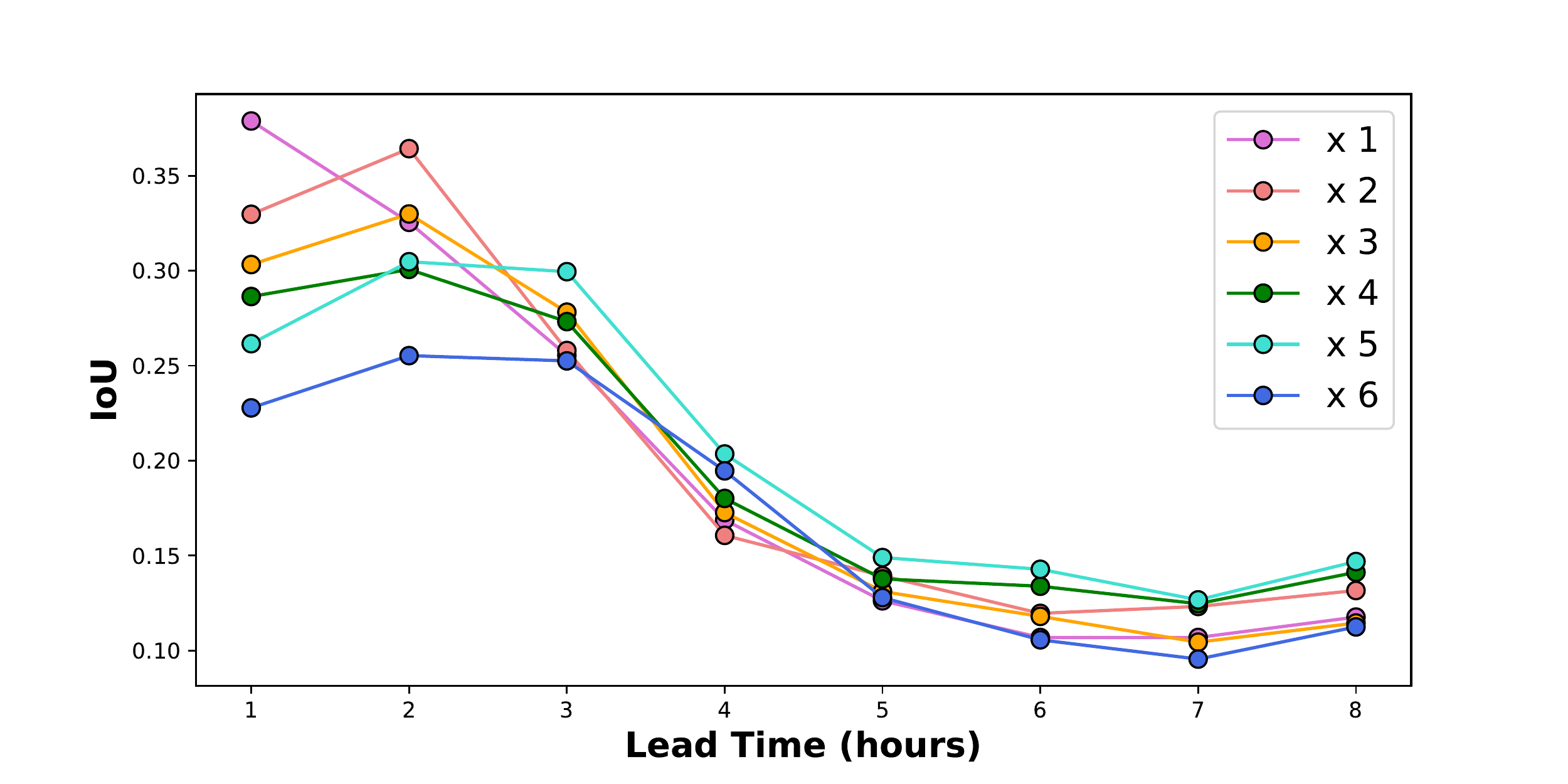}
\vspace{-0.5cm}
\caption{Comparison of crop size performances over lead times. The crop size $\times N$ indicates how many times larger the crop region is than the target region.
}
\vspace{0.05cm}
\label{fig:cropsize}
\end{minipage}
\hspace{0.03\textwidth}
\begin{minipage}[b]{0.32\textwidth}
\centering

  \begin{tabular}{cc}
    \toprule
    
    Crop size &Test Score\\
    \midrule
    $\times1$ &0.2380\\
    $\times2$&  0.2623\\
    $\times3$&  \textbf{0.2685}\\
    $\times4$& 0.2600\\
   $\times5$&  0.2454\\
    $\times6$ (non-crop)&  0.2403\\
    \bottomrule
\end{tabular}
\vspace{0.5cm}
\captionof{table}{The test leaderboard score corresponding to the crop size for stage 2. Best in bold }
\label{tab:crop}
\vspace{0.05cm}
\end{minipage}
    \end{figure}

 
\Cref{fig:cropsize} shows IoU performance over lead time according to a crop size. 
 Crop size $\times N$ represents the size of the cropped region that is $N$ times larger than the target region.
 A crop size $\times 1$, for example, indicates that we crop only the 42 $\times$42 pixel area in the center of the input frame, which is the same size with the target region (the 42 $\times$42 pixels).
 Moreover, a crop size ×6 is a non-crop scenario since it crops an equal size of the input frame (the 252 × 252 pixels). 
\clearpage
In the case of only the small target area without surrounding context is used for training(pink line, crop size $\times1$, in \cref{fig:cropsize}), the performance within a short term period is the best among the various crop size, but the performance rapidly degrades over time.
Conversely, when looking at wider areas (see blue $\times 6$ or turquoise $\times 5$ line in \cref{fig:cropsize}), short term prediction performance becomes lower, but performances of long term prediction is higher compared to the pink line.
These results indicate that rich surrounding information improves long-term predictive performance, however there is a trade-off with short-term predictive performance.
Based on these observations, we chose crop size $\times$ 3 for our method.
\Cref{tab:crop} summarizes the test score results.

\begin{figure}[ht]
      \centering
      \resizebox{13.5cm}{!}{
      \includegraphics{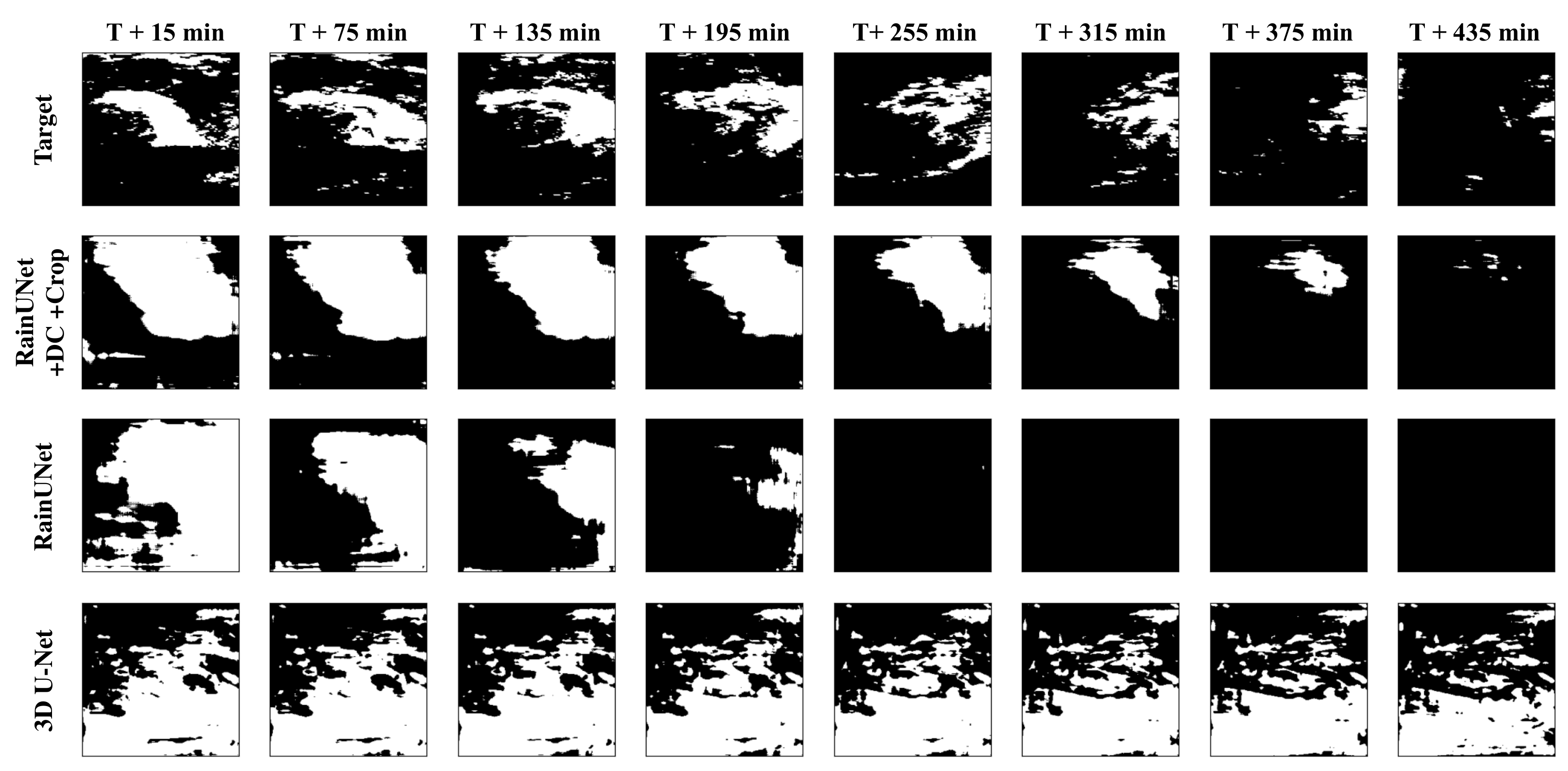}
      }
      \caption{Comparison with a baseline. We visualize the predictions at one-hour intervals.}
\label{fig:prediction_over_time}      
\end{figure}
\begin{figure}[ht]
      \centering
      \resizebox{13.5cm}{!}{
      \includegraphics{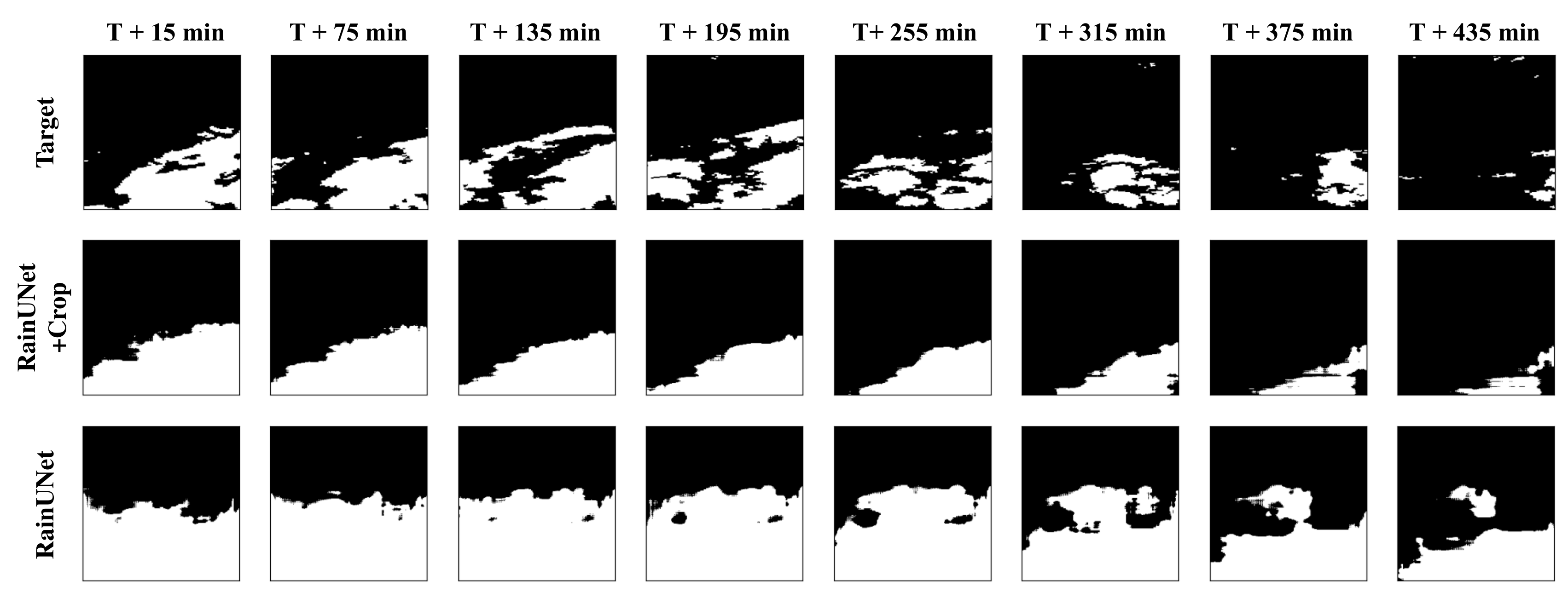}
      }
      \caption{Comparison of predicted results according to whether the center crop was utilized. We visualize the predicted results at one hour intervals.}
      
\label{fig:prediction_over_time_crop}
\end{figure}
For qualitative results, we visualize the output of 3D U-Net and RainUNet in \cref{fig:prediction_over_time}. 
The outputs of 3D U-Net are similar regardless of time; however, our method captures the overall change in rain rates over time, demonstrating our method’s superiority in capturing temporal relationships.

More specifically, the center crop has a crucial role in our model, as shown in \cref{fig:prediction_over_time_crop}.
Without the center crop, a slight output variation occurs over time, but including the center crop reveals a more detailed temporal relationship.
This shows that removing unnecessary information helps the model learn well. 

\clearpage
\begin{table}[ht!]
\vspace{1cm}
  \caption{
  Validation IoU and scores from Weather4cast 2022 competition.
  Best in bold.}
  \vspace{0.1cm}
  \label{tab:score_table}
  \centering
    
  \begin{tabular}{c|c c c c}
    \toprule
    Model & Valid IoU & Test score & Core heldout score & Transfer heldout score \\
    \midrule
    3D U-Net & 0.2134 & 0.2256 &  0.2551 & 0.1987 \\
    RainUNet & \textbf{0.2580} & \textbf{0.2685} & \textbf{0.2871} & \textbf{0.2357} \\
    \bottomrule
  \end{tabular} 
  
  \label{tab:final}
\end{table}

Finally, we report the leaderboard score of our solution as shown in \cref{tab:score_table}.
We note that our best evaluation results from the held-out test set are 0.287 in the core challenge task and 0.236 in the transfer challenge task. The baseline score is based on what was submitted by the organizer. 




\section{Conclusion}
This paper has proposed RainUNet, a simple yet novel weather forecasting model, as a solution to the Weather4cast challenge. 
Particularly, RainUNet is a hierarchical U-shaped network with TS blocks that uses factorized 3D convolution, which combines the advantages of convolution and self-attention.
Also, we have introduced a novel 3D Convolution, called factorized 3D convolution (F3DConv), which is a key component of our RainUNet.
F3DConv decomposes 3D convolution into spatial and time axes and forcing the 3D convolution into discrete spatial and temporal components to not only increase interframe correlation.
Additionally, we have utilized two data processing policies to debias and generalize future prediction, further enabling the capabilities of RainUNet. 
 
 In comprehensive ablation studies, we demonstrate the effectiveness of our proposed methods.
The proposed model shows a higher performance than the baseline model in the experimental settings of the Weather4cast 2022 core challenge.

Since the proposed model does not use additional location and time information of input data, it could be applicable even when such information is not provided.
Overall, RainUNet obtains precise weather forecasting results in the experimental environment of the Weather4cast 2022 challenge.
Furthermore, we expect that the model will be expandable for general video prediction tasks as well as weather forecasting since our proposed TS block can exploit intraframe and interframe correlation among video sequences using surrounding information.

\newpage



\begin{thebibliography}{17}
\providecommand{\natexlab}[1]{#1}
\providecommand{\url}[1]{\texttt{#1}}
\expandafter\ifx\csname urlstyle\endcsname\relax
  \providecommand{\doi}[1]{doi: #1}\else
  \providecommand{\doi}{doi: \begingroup \urlstyle{rm}\Url}\fi

\bibitem[Bai et~al.(2022)Bai, Sun, Zhang, Song, and Chen]{bai2022rainformer}
Cong Bai, Feng Sun, Jinglin Zhang, Yi~Song, and Shengyong Chen.
\newblock Rainformer: Features extraction balanced network for radar-based
  precipitation nowcasting.
\newblock \emph{IEEE Geoscience and Remote Sensing Letters}, 19:\penalty0 1--5,
  2022.

\bibitem[{\c{C}}i{\c{c}}ek et~al.(2016){\c{C}}i{\c{c}}ek, Abdulkadir, Lienkamp,
  Brox, and Ronneberger]{cciccek20163d}
{\"O}zg{\"u}n {\c{C}}i{\c{c}}ek, Ahmed Abdulkadir, Soeren~S Lienkamp, Thomas
  Brox, and Olaf Ronneberger.
\newblock {3D} {U-N}et: learning dense volumetric segmentation from sparse
  annotation.
\newblock In \emph{International conference on medical image computing and
  computer-assisted intervention}, pages 424--432. Springer, 2016.

\bibitem[Gruca et~al.(2021)Gruca, Herruzo, R\'{\i}podas, Kucik, Briese, Kopp,
  Hochreiter, Ghamisi, and Kreil]{10.1145/3459637.3482044}
Aleksandra Gruca, Pedro Herruzo, Pilar R\'{\i}podas, Andrzej Kucik, Christian
  Briese, Michael~K. Kopp, Sepp Hochreiter, Pedram Ghamisi, and David~P. Kreil.
\newblock \emph{CDCEO'21 - First Workshop on Complex Data Challenges in Earth
  Observation}, page 4878–4879.
\newblock Association for Computing Machinery, New York, NY, USA, 2021.
\newblock ISBN 9781450384469.
\newblock URL \url{https://doi.org/10.1145/3459637.3482044}.

\bibitem[Guo et~al.(2022)Guo, Lu, Liu, Cheng, and Hu]{guo2022visual}
Meng-Hao Guo, Cheng-Ze Lu, Zheng-Ning Liu, Ming-Ming Cheng, and Shi-Min Hu.
\newblock Visual attention network.
\newblock \emph{arXiv preprint arXiv:2202.09741}, 2022.

\bibitem[Herruzo et~al.(2021)Herruzo, Gruca, Lliso, Calbet, Rípodas,
  Hochreiter, Kopp, and Kreil]{9672063}
Pedro Herruzo, Aleksandra Gruca, Llorenç Lliso, Xavier Calbet, Pilar Rípodas,
  Sepp Hochreiter, Michael Kopp, and David~P. Kreil.
\newblock High-resolution multi-channel weather forecasting – first insights
  on transfer learning from the weather4cast competitions 2021.
\newblock In \emph{2021 IEEE International Conference on Big Data (Big Data)},
  pages 5750--5757, 2021.
\newblock \doi{10.1109/BigData52589.2021.9672063}.

\bibitem[Izmailov et~al.(2018)Izmailov, Podoprikhin, Garipov, Vetrov, and
  Wilson]{izmailov2018averaging}
Pavel Izmailov, Dmitrii Podoprikhin, Timur Garipov, Dmitry Vetrov, and
  Andrew~Gordon Wilson.
\newblock Averaging weights leads to wider optima and better generalization.
\newblock \emph{arXiv preprint arXiv:1803.05407}, 2018.

\bibitem[Klocek et~al.(2021)Klocek, Dong, Dixon, Kanengoni, Kazmi, Luferenko,
  Lv, Sharma, Weyn, and Xiang]{klocek2021ms}
Sylwester Klocek, Haiyu Dong, Matthew Dixon, Panashe Kanengoni, Najeeb Kazmi,
  Pete Luferenko, Zhongjian Lv, Shikhar Sharma, Jonathan Weyn, and Siqi Xiang.
\newblock {MS}-nowcasting: Operational precipitation nowcasting with
  {C}onvolutional {LSTM}s at microsoft weather.
\newblock \emph{arXiv preprint arXiv:2111.09954}, 2021.

\bibitem[Loshchilov and Hutter(2017)]{loshchilov2017decoupled}
Ilya Loshchilov and Frank Hutter.
\newblock Decoupled weight decay regularization.
\newblock \emph{arXiv preprint arXiv:1711.05101}, 2017.

\bibitem[Qiu et~al.(2017)Qiu, Yao, and Mei]{qiu2017learning}
Zhaofan Qiu, Ting Yao, and Tao Mei.
\newblock Learning spatio-temporal representation with pseudo-3d residual
  networks.
\newblock In \emph{proceedings of the IEEE International Conference on Computer
  Vision}, pages 5533--5541, 2017.

\bibitem[Qiu et~al.(2019)Qiu, Yao, Ngo, Tian, and Mei]{qiu2019learning}
Zhaofan Qiu, Ting Yao, Chong-Wah Ngo, Xinmei Tian, and Tao Mei.
\newblock Learning spatio-temporal representation with local and global
  diffusion.
\newblock In \emph{Proceedings of the IEEE/CVF Conference on Computer Vision
  and Pattern Recognition}, pages 12056--12065, 2019.

\bibitem[Qiu et~al.(2022)Qiu, Yao, Ngo, and Mei]{qiu2022optimization}
Zhaofan Qiu, Ting Yao, Chong-Wah Ngo, and Tao Mei.
\newblock Optimization planning for 3d convnets.
\newblock \emph{arXiv preprint arXiv:2201.04021}, 2022.

\bibitem[Sudre et~al.(2017)Sudre, Li, Vercauteren, Ourselin, and
  Jorge~Cardoso]{sudre2017generalised}
Carole~H Sudre, Wenqi Li, Tom Vercauteren, Sebastien Ourselin, and
  M~Jorge~Cardoso.
\newblock Generalised dice overlap as a deep learning loss function for highly
  unbalanced segmentations.
\newblock In \emph{Deep learning in medical image analysis and multimodal
  learning for clinical decision support}, pages 240--248. Springer, 2017.

\bibitem[Tran et~al.(2018)Tran, Wang, Torresani, Ray, LeCun, and
  Paluri]{tran2018closer}
Du~Tran, Heng Wang, Lorenzo Torresani, Jamie Ray, Yann LeCun, and Manohar
  Paluri.
\newblock A closer look at spatiotemporal convolutions for action recognition.
\newblock In \emph{Proceedings of the IEEE conference on Computer Vision and
  Pattern Recognition}, pages 6450--6459, 2018.

\bibitem[Trebing et~al.(2021)Trebing, Stanczyk, and
  Mehrkanoon]{trebing2021smaat}
Kevin Trebing, Tomasz Stanczyk, and Siamak Mehrkanoon.
\newblock {S}ma{A}t-{UN}et: Precipitation nowcasting using a small
  attention-{UN}et architecture.
\newblock \emph{Pattern Recognition Letters}, 145:\penalty0 178--186, 2021.

\bibitem[Veillette et~al.(2020)Veillette, Samsi, and
  Mattioli]{veillette2020sevir}
Mark Veillette, Siddharth Samsi, and Chris Mattioli.
\newblock Sevir: A storm event imagery dataset for deep learning applications
  in radar and satellite meteorology.
\newblock \emph{Advances in Neural Information Processing Systems},
  33:\penalty0 22009--22019, 2020.

\bibitem[Woo and Wong(2017)]{woo2017operational}
Wang-chun Woo and Wai-kin Wong.
\newblock Operational application of optical flow techniques to radar-based
  rainfall nowcasting.
\newblock \emph{Atmosphere}, 8\penalty0 (3):\penalty0 48, 2017.

\bibitem[Wu and He(2018)]{wu2018group}
Yuxin Wu and Kaiming He.
\newblock Group normalization.
\newblock In \emph{Proceedings of the European conference on computer vision
  (ECCV)}, pages 3--19, 2018.

\end{thebibliography}
\bibliographystyle{plainnat}






\end{document}